\documentclass[10pt,twocolumn,letterpaper]{article}
\usepackage[accsupp]{axessibility}
\usepackage{iccv}
\usepackage{times}
\usepackage{epsfig}
\usepackage{graphicx}
\usepackage{amsmath}
\usepackage{amssymb}

\usepackage{color}
\usepackage{xcolor}
\usepackage{booktabs}
\usepackage{pdfpages}
\usepackage[font=small]{caption}

\iccvfinalcopy 
\definecolor{citecolor}{HTML}{0071bc}

\usepackage[pagebackref=true,breaklinks=true,letterpaper=true,colorlinks,bookmarks=false]{hyperref}
\def\iccvPaperID{6423} 

\ificcvfinal\fi

\usepackage[capitalize]{cleveref}
\crefname{section}{Sec.}{Secs.}
\Crefname{section}{Section}{Sections}
\Crefname{table}{Table}{Tables}
\crefname{table}{Tab.}{Tabs.}

\newcommand{\dino}{f}

\definecolor{mypink}{RGB}{166, 127, 124}
\definecolor{myblue}{RGB}{97, 121, 157}

\newcommand\blfootnote[1]{%
  \begingroup
  \renewcommand\thefootnote{}\footnote{#1}%
  \addtocounter{footnote}{-1}%
  \endgroup
}

\begin{document}

\title{3DMiner: Discovering Shapes from Large-Scale Unannotated Image Datasets}

\author{Ta-Ying Cheng$^{1}\footnotemark$ \ \ Matheus Gadelha$^2$ \ \ S\"{o}ren Pirk$^2$ \ \ Thibault Groueix$^2$ \\ \ \ Radom\'{i}r M\u{e}ch$^2$ \ \ Andrew Markham$^1$ \ \ Niki Trigoni$^1$
\vspace{0.2cm}
\\\hspace{-0.1cm}$^1$University of Oxford \ \ $^2$Adobe Research
\vspace{0.08cm}
\\
}



\twocolumn[\maketitle
\begin{center}
  \includegraphics[width=\linewidth]{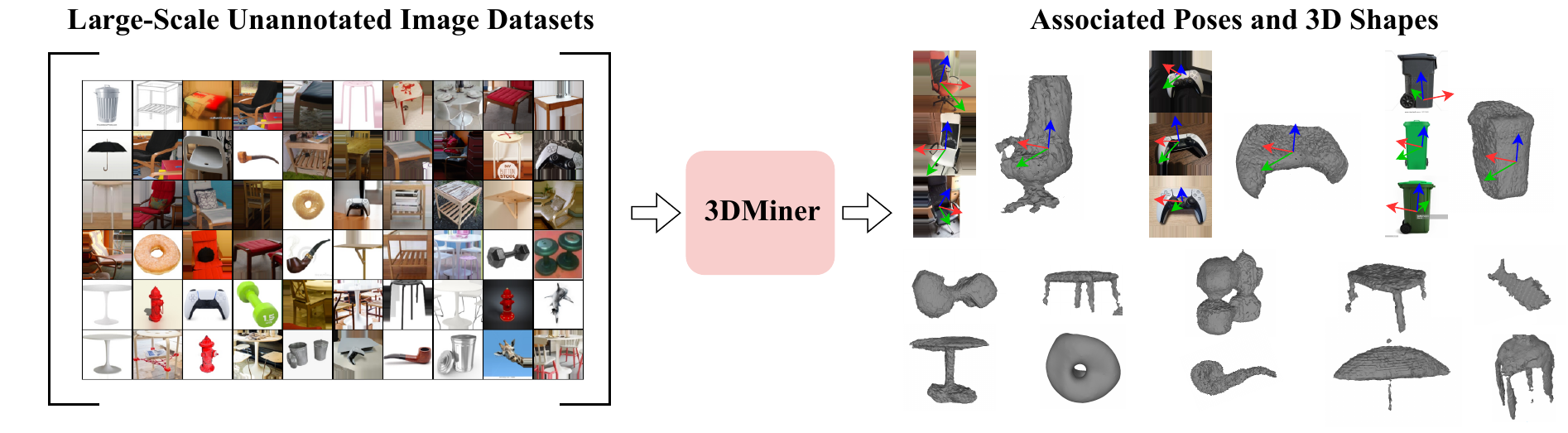}
\end{center}
\vspace{-9mm}
\captionof{figure}{\textbf{Overview}. We present 3DMiner, a scalable framework designed to obtain associating poses and reconstruct shapes from \emph{diverse and realistic} sets of images \emph{without any 3D data, pose annotation, camera information, or keypoints}.
}

\label{fig:teaser}
\bigbreak]

\label{fig:teaser}


\begin{abstract}
We present 3DMiner -- a pipeline for mining 3D shapes from challenging large-scale unannotated image datasets. 
Unlike other unsupervised 3D reconstruction methods, we assume that, within a large-enough dataset, there must exist images of objects with similar shapes but varying backgrounds, textures, and viewpoints. 
Our approach leverages the recent advances in learning self-supervised image representations to cluster images with geometrically similar shapes and find common image correspondences between them. 
We then exploit these correspondences to obtain rough camera estimates as initialization for bundle-adjustment.
Finally, for every image cluster, we apply a progressive bundle-adjusting reconstruction method to learn a neural occupancy field representing the underlying shape.
We show that this procedure is robust to several types of errors introduced in previous steps (e.g., wrong camera poses, images containing dissimilar shapes, etc.), allowing us to obtain shape and pose annotations for images in-the-wild.
When using images from Pix3D chairs, our method is capable of producing significantly better results than state-of-the-art unsupervised 3D
reconstruction techniques, both quantitatively and qualitatively.
Furthermore, we show how 3DMiner can be applied to in-the-wild data by reconstructing shapes present in images from the LAION-5B dataset. Project Page: \href{https://ttchengab.github.io/3dminerOfficial}{https://ttchengab.github.io/3dminerOfficial}.
\end{abstract}


\blfootnote{\textsuperscript{*}Work was partially done during internship at Adobe Research.}
\vspace{-50pt}
\section{Introduction}
\label{sec:intro}

Learning-based systems that try to reason about 3D geometry from images suffer from 
a fundamental limitation: the amount of available 3D data.
Despite recent advances in capturing the world tridimensionally, the biggest image datasets still contain many orders of magnitude more
data points than their 3D counterparts.
In practice, the sheer quantity of images ends up capturing a much richer visual vocabulary -- different textures, illuminations, shapes, environments,
types of objects and relationships between them.
Therefore, developing techniques that can leverage this information is the key to general,
high-performing 3D reconstruction algorithms.
Unfortunately, the abundance and variety of visual data in image datasets leads to great complexity when trying to extract 3D information.
Consider a very simple dataset consisting of multiple images of the same object in the same environment taken from
different camera viewpoints.
Extracting 3D information is not entirely trivial, but potentially doable through stucture-from-motion approaches.
However, if we increase the complexity of this dataset by adding images of the object in different environments, different materials and with slight shape variations, structure-from-motion techniques will fail.
The complexity only increases when we consider all the possible image permutations that one can find online:
a myriad of object types, occluded objects, partial observations, non-photorealistic imagery and so on. 
How can we extract 3D information from such complicated image datasets?

Various image-to-3D approaches have tried to tame the visual complexity in big image datasets. 
They usually employ different amounts of manual image
annotations; \textit{i.e.,} object poses, masks, keypoints, part segmentations, and so on. 
These techniques use models that are trained to disentangle 3D geometry from various other factors
while trying to reconstruct the original image and its annotations.
Due to the ill-posed nature of the single-view reconstruction problem, training these models is very hard and multiple
regularizations are necessary.
When presented with more challenging datasets, with real images, even if they only depict a single object type (e.g., Pix3D chairs),
the best models fail to produce reasonable results.

In this work we aim to extract 3D shapes from image datasets in a completely different way.
We call our approach 3DMiner.
Given a very large set of images, our initial goal is to separate them in groups containing similar shapes. 
Within these groups, we estimate robust pairwise image correspondences that will give us a good idea about the relative pose of the objects in the images. 
Using this information, we can estimate the underlying 3D geometry in every image group, effectively treating the
single-view reconstruction problem as a noisy multi-view one.
Unfortunately, this is not a straightforward structure-from-motion setup -- within the same group, the objects are similar
but not exactly the same; they have different colors, backgrounds, and even slightly different geometry.
To circumvent this issue, we adopt modern reconstruction techniques based on neural fields.
These representations give us the ability to train parametric occupancy fields through gradient descent
while refining camera poses, intrinsics, and more importantly, giving us a proxy for the quality of the recovered
3D shape -- the image reconstruction loss. 
Ultimately, the entire pipeline provides association between the shape and poses across images in-the-wild for arbritrary categories -- a task for which many datasets rely on manual annotations.

Instead of having a single hard-to-train model extracting shapes from images, we opt for dissecting the problem and breaking
it in pieces that can be tackled with well-studied tools.
At the heart of our approach are recent advances in learning representations from images in an unsupervised manner. 
Those techniques allow us to identify image associations and to find common features more conveniently, 
establishing robust correspondences and ignoring nuisance factors like different backgrounds, illumination and small shape variations. 
As a notable example, the recent DINO-ViT, trained through self-supervision on ImageNet data, has shown remarkable ability 
to distinguish foregrounds, perform part segmentation, and generate common keypoints.
More importantly: further improvements in image representation learning can be \emph{immediately} incorporated into our method -- no model needs to be retrained or fine-tuned.
The same can also be applied to advances in neural fields.
Once better methods for optimizing occupancy fields are developed, they can be directly plugged into our pipeline.

We refer to our method as unsupervised, meaning that it does not require 3D data to perform 3D reconstruction. Thus, using other features/outputs from models trained without 3D data do not alter the unsupervised 3D reconstruction setup.
We do, however, restrain from using keypoints/pose estimators as they limit our approach to category-specific conventions.

We demonstrate the capabilities of our approach in two empirical studies using Pix3D and LAION-5B. 
For Pix3D, we use our 3D mining pipeline as a way of generating 3D for each image in the dataset and directly
compare it against state-of-the-art single-view reconstruction approaches that do not use 3D information
during training.
Our experiments show that, differently from the other approaches, 3DMiner is capable of generating reasonable 3D shapes and displays a relative F-score improvement of \textbf{80\%} over the state-of-the-art.
In order to showcase the versatility of our approach we also ran 3DMiner on subsets of images from LAION-5B gathered using various short text prompts.
Despite the challenges presented by the diversity in this dataset, 3DMiner is still capable of finding reasonable 3D representations across multiple categories.

In summary, our contributions are: 
\emph{(i)} a new problem formulation on mining 3D shapes from large-scale web-retrieved images without any priors or annotations; 
\emph{(ii)} an end-to-end pipeline to cluster, estimate pose, and generate neural 3D representations from unannotated image datasets without any 3D ground truths; 
\emph{(iii)} a detailed empirical study to showcase the superiority of our method on challenging datasets and robustness under categories where no previous ground-truth reconstructions exist.

\begin{figure*}[t]
\centering
\includegraphics[width=\linewidth]{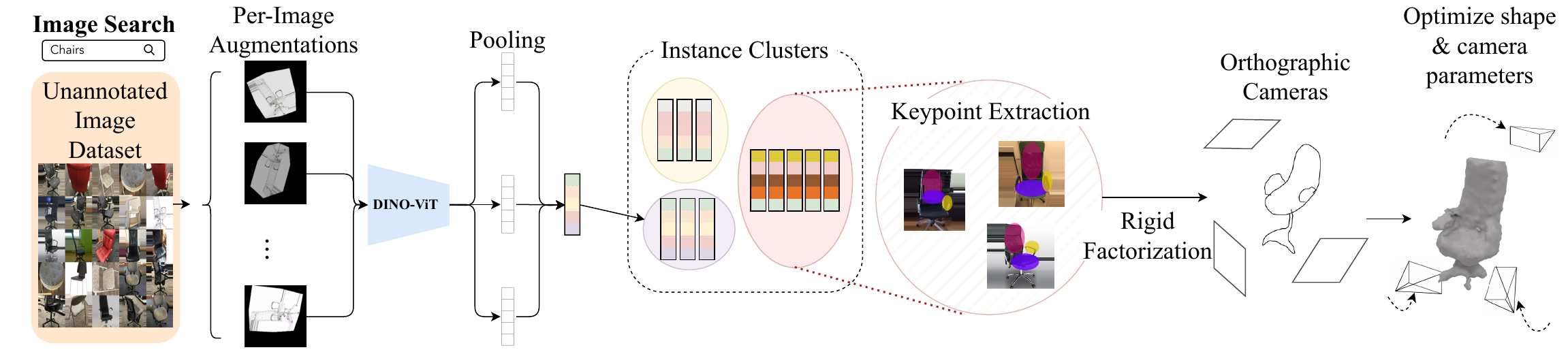}
\caption{
\label{fig:pipeline}
\textbf{3DMiner pipeline.} Our method starts by grouping images that depict similar 3D shapes, regardless of the texture of the shape, the camera view-point or the background. To do so, we perturb each image  with various transformations (e.g. color jittering, perspective and rotation) and we pool their DINO-ViT features to create a robust image embedding.
We cluster images by running agglomerative clustering on the embeddings.
Within each cluster, we find key point correspondences using dense DINO-ViT features. We feed those corresponding keypoints to a Structure from Motion algorithm (rigid factorization) to get coarse orthographic camera estimations.
Finally, we jointly refine the camera parameters and learn an occupancy field to get the final shape.
}
\end{figure*}

\section{Related Work}
\paragraph{Multi-view reconstruction,} or
  \textit{{Structure from Motion} (SfM)}, assumes a set of images of a stationary scene and jointly reconstructs a 3D point cloud of the scene along with the camera parameters of each image.
  Since the seminal work of Longuet-Higgins~\cite{longuet}, SfM has been extensively researched~\cite{tomasikanade,btriggs,snavely2006photo,agarwal2009building}.
  Before optimization, the camera parameters are typically initialized by either applying RANSAC~\cite{ransac} or matrix factorization on pairs of keypoints. Classical approaches to keypoint matching detect candidate matches in each image  and embed them into a descriptive feature space.
  More recently, classical descriptors~\cite{sift} have been outperformed by learned approaches~\cite{detone2018superpoint,sarlin20superglue}. 
  Interestingly, features from self-supervised approaches, like DINO-ViT~\cite{caron2021emerging}, have proven to be robust in a wide range of downstream tasks, including keypoint matching~\cite{amir2021deep}. 
    
    More recently, Mildenhall \textit{et al.}~\cite{mildenhall2020nerf} demonstrated photo-realistic performance on the task of novel-view synthesis using differentiable volume rendering of neural radiance fields. NeRFs~\cite{mildenhall2020nerf} initially assumed multiple views of the same stationary scene with known camera poses and several extensions are relevant to our work. NeRF-W~\cite{nerfw} tries to generalize NeRF to more diverse conditions, in particular transient occluders and illumination changes. Several approaches address 3D reconstruction instead of novel view synthesis and optimize directly sign-distance fields~\cite{zhang2021learning, yariv2020multiview, neus, volsdf} or occupancy fields~\cite{dvr, unisurf}. BARF~\cite{lin2021barf} assumes coarse camera poses, and jointly performs bundle-adjustment and optimizes the radiance field. Like BARF, we jointly optimize an occupancy field and perform a bundle-adjustment.
    
    However, SfM approaches, NeRF, and its variants assume photometric consistency between the different views of the scene. On the contrary, our approach, 3DMiner, does not rely on this assumption and reconstructs 3D shapes from image clusters containing roughly the same geometric shape with different textures and backgrounds.
    We draw inspiration from the classical SfM pipeline: we use  DINO-ViT~\cite{caron2021emerging} features to estimate corresponding keypoints across images with different textures and backgrounds, and use those keypoints to estimate coarse camera poses. Similar to BARF~\cite{lin2021barf} and space carving~\cite{spacecarving}, we then jointly reconstruct an occupancy field and perform bundle-adjustments, using a silhouette loss.
    Silhouettes can be estimated using the saliency from DINO-ViT features and further refined using saliency segmentation models.
    
\noindent\textbf{Learning shapes with 3D data.}
Several approaches leverage existing datasets of 3D shapes, \textit{e.g.,} ShapeNet~\cite{chang2015shapenet}, ModelNet~\cite{modelnet40}, Pix3D~\cite{sun2018pix3d}, to learn 3D reconstructions. Their most distinctive feature is usually the choice of 3D representation. 3D geometry can be represented via voxel grids~\cite{3dr2n2,xie2020pix2vox++,cheng2022pose}, point clouds~\cite{point_set_gen, mandikal20183dlmnet}, meshes~\cite{atlasnet,pixel2mesh,8491020}, or implicit functions~\cite{mescheder2019occupancy,im-net,deepsdf,yu2021pixelnerf}.
A common challenge for this type of approach is to generalize beyond the limited number of categories they are trained on. On the contrary, 3DMiner seeks to leverage large-scale in-the-wild 2D datasets for their potential to cover a wider range of objects and not limited to specific categories.

\noindent\textbf{Learning shapes without 3D data.} 
A large corpus of work in the \textit{Single-Image Reconstruction (SVR)} community focuses on directly learning 3D from 2D images,
without any 3D annotation. Several approaches supervise SVR with multiple views of the same scenes using differentiable volumetric rendering~\cite{drcTulsiani17,perspective_transformer_net,rezende,dvr} or differentiable mesh renderers~\cite{kato2018renderer,softras,chen2019_dibr}. Notably, ~\cite{mvcTulsiani18, insafutdinov2018pointclouds} do not assume known camera poses but estimate them jointly.

 
To overcome the ill-posed nature of the problem, several forms of priors have been explored, including  keypoints correspondences~\cite{kar2015category, reconstructingpascalvoc, drcTulsiani17, cmrKanazawa18, tars3d, lin2020sdfsrn, henderson20cvpr}, silhouette losses
 \cite{cmrKanazawa18,u-cmr,imr,smr,radar,ye2021shelf}, shape templates~\cite{u-cmr, imr}, different forms of symmetry~\cite{henderson20cvpr, u-cmr, cmrKanazawa18, umr2020, smr} including rotation symmetries~\cite{radar}.  Several approaches leverage off-the-shelf 2D networks~\cite{umr2020,ye2021shelf} or generative adversarial techniques~\cite{henzler2019platonicgan,prgan,kato2019vpl,pavllo2020convmesh,ye2021shelf}. 
 Particularly relevant to us are Unicorn~\cite{monnier2022share} and SMR~\cite{smr}. In SMR, Hu \textit{et al.}~\cite{smr} use self-supervised learning to learn 3D from 2D images. Their method requires only images and their corresponding silhouettes. In Unicorn,  Monnier \textit{et al.}~\cite{monnier2022share} propose a progressive conditioning strategy, only assuming that images in the training set belong to the same category. We compare the performances of 3DMiner against these recent techniques. Note however, that while SMR and Unicorn tackle SVR, 3DMiner is not intended to be used for SVR but rather to automatically mine 3D content from large 2D collections of images.

Despite the gradual improvements in accuracy and the removal of label requirements, 
all of these methods are still not capable of yielding good results in more challenging datasets. 
Most comparisons mainly exist on synthetic data~\cite{chang2015shapenet} 
or very constrained real-world images~\cite{wah2011caltech},
where the background and foreground are very distinct and not many different shapes can be found. 
Previous approaches cannot handle data such as Pix3D \cite{sun2018pix3d} 
(unless trained with additional datasets), not to mention in-the-wild datasets such as LAION-5B \cite{schuhmannlaion}.
The challenge comes from the fact that training the network to
generate reasonable 3D geometry by using reprojection losses is very hard -- it requires a lot of regularizations that lead to overly smooth geometry, otherwise yielding degenerate solutions.
In contrast, our approach adopts a pipeline where images are grouped by shape similarity and the reconstruction
in each group happens independently.
This endows 3DMiner with the ability to not only tackle more challenging data but also leads to a system
that readily benefits from progress in relevant subproblems 
(\textit{e.g.,} image representation learning, landmark detection, pose estimation, neural field reconstruction) 
without requiring any retraining or fine-tuning.

\section{Method}


\paragraph{Overview.}

Our goal is to mine 3D shapes from large-scale in-the-wild image datasets. 
We assume that, given a large-enough dataset, there must exist several images of very similar shapes once ignored the differences in terms of backgrounds, textures, viewpoints, and lightning conditions. 
When the images containing similar shapes are grouped, we extract pairwise image correspondences to estimate
orthographic camera poses for each image.
Finally, this information will be used in a neural occupancy field optimization procedure that 
recovers shape and refines perspective camera poses.
Figure~\ref{fig:pipeline} shows an overview of our method.
In the following subsections we describe each step of this pipeline in detail.


  \begin{table*}[t!]
    \centering
    \resizebox{0.6\linewidth}{!}{

    \begin{tabular}{lcc|cc|cc|cc}
    \toprule
    & & & \multicolumn{6}{c}{\textbf{Threshold}} \\
    & \multicolumn{2}{c|}{} & \multicolumn{2}{c|}{0.5} & \multicolumn{2}{c|}{0.4} & \multicolumn{2}{c}{0.3} \\
    & \textbf{CD $\downarrow$}& \textbf{F1 $\uparrow$} & \textbf{CD $\downarrow$}& \textbf{F1 $\uparrow$} & \textbf{CD $\downarrow$}& \textbf{F1 $\uparrow$} & \textbf{CD $\downarrow$}& \textbf{F1 $\uparrow$}\\
    \midrule
    SMR \cite{smr}      & 0.192 & 0.130 & 0.189 & 0.131& 0.188 & 0.132 & 0.267 & 0.110\\
    Unicorn \cite{monnier2022share} & 0.263 & 0.102  & 0.259 & 0.106& 0.266& 0.105 & 0.154 & 0.160 \\
    3DMiner (Ours)      &\textbf{0.130} & \textbf{0.234} &  \textbf{0.125}  &  \textbf{0.244}&  \textbf{0.116}  &  \textbf{0.263} &   \textbf{0.095} &  \textbf{0.307} \\    
     
    \bottomrule
    \end{tabular}
    }
    \caption{\textbf{Comparisons on Pix3D Chairs.} We align the meshes to their ground truths via Coherent Point Drift~\cite{coherentpointdrift} and compute the Chamfer Distance (CD) and F1 Score (F1). We select three subsets of images using a threshold on the reprojection error within each cluster.  On the full Pix3D chairs (left two columns), 3DMiner improves by 33\% the CD and 10 points the F1, and the performance increases significantly when we use our selection criterion (see Section~\ref{sec:pix3d} for more discussion).
    }
    \label{tab:pix3d}
    \end{table*}

\subsection{Clustering similar shapes}
\label{sec:cluster}
We aim to find clusters of similar shapes, irrespective of their texture, background and illumination conditions. 
Caron \textit{et al.}~\cite{caron2021emerging} showed that Vision Transformers (ViT) trained with self-supervision learned powerful features for classification tasks as well as semantic segmentation tasks.
We take advantage of these  models to perform a clustering of images containing similar shapes. 
Contrary to 3D reconstruction approaches trained with 3D supervision on selected object categories,  DINO-ViT has been trained on ImageNet~\cite{ILSVRC15} and therefore is more robust when applied to in-the-wild imagery.

To make the clustering invariant to texture, background and the camera viewpoint, we  propose a simple augmentation method. 
Formally, given an image $I$, we obtain a set of augmented DINO-ViT features $S_I = \{\dino(A_1(I)), ...,\dino(A_n(I))\}$, 
where $A_i$ refers to a random set of augmentations involving color jittering, 
image rotation, and perspective transformation, and $\dino(\cdot)$ is the pretrained DINO-ViT outputing a per-image global feature. 
The final feature of the image $z_I$ is then computed as:
\begin{equation}
    z_I = [\max (S_I), \mathrm{mean}(S_I), \min(S_I)].
    \label{eqn:inf_aug}
\end{equation}

Color jittering helps to make the resulting feature more invariant to texture and illumination conditions. Rotation and perspective transformations mimic a  change of viewpoints in 3D. These augmentations encourage the features to be dependent on the geometry itself.

Using the feature $z_I$ from Equation \ref{eqn:inf_aug}, we perform a bottom-up agglomerative clustering. 
 Section \ref{sec:inf-augmentation} presents an analysis of our augmentation scheme on the Pix3D datase and Figure~\ref{fig:laion} shows examples of clusters from the LAION-5B dataset.

\subsection{Coarse orthographic pose estimation}
At this point, we assume we have a cluster of unannotated images of the same geometric shape and seek to estimate a coarse camera pose for each image in the cluster.
 Over a decade ago, Marques et al. proposed a matrix factorization technique to estimate orthographic poses from a sequence of images and corresponding keypoints~\cite{rigid_factorization}. 
 Their method is robust to some amount of noise in the keypoints and does not require that all images contain all keypoints. Perhaps for this reason it has been widely adopted in datasets such as CUB-200 and Pascal VOC where manually labelled keypoints are available~\cite{cmrKanazawa18,reconstructingpascalvoc,kar2015category}. 
 Thus, we will leverage this classical technique to estimate initial camera poses, but, differently from previous approaches,
 our keypoint correspondences are established using image representations learned through self-supervision.
 More specifically, we extract image features using the DINO-ViT model (same used for clustering).

Within a cluster, we adopt an approach inspired by part-segmentation method from~\cite{amir2021deep}. 
We extract features at different layers of the ViT for all spatial locations in all images, 
run k-means on this set of features, and select segments that are salients and common to most images using a voting strategy. 
We compute the bounding boxes of each segment in each image, and use its center as a keypoint. 
We show in the supplementary material a visualization of the estimated keypoints for a multiple clusters.

Equipped with estimated keypoints within a cluster, we then perform rigid factorisation through SVD and Stiefel manifold projections following~\cite{rigid_factorization} to obtain an orthographic camera for every image $I$ in the format:
\begin{equation}
    p_{2d} = M \cdot p_{3d} + t,
    \label{eqn:orthographic}
\end{equation}
where $M \in \mathbb{R}^{2\times 3}$ and  $t \in \mathbb{R}^{2\times 1}$ are the orthographic motion and translation matrix to project a 3D point $p_{3d}$ to a 2D point $p_{2d}$ on the image plane of $I$. 

\subsection{Bundle-adjusting neural occupancy field}

From now on, we assume we have a set of images with the same geometric shape \emph{and} a rough pose for every image within each cluster. 
Directly applying a variation of NeRF~\cite{mildenhall2020nerf} is not a viable approach as there exist no photo-consistency due to the difference in textures and backgrounds. 
Therefore, we propose to use silhouettes instead of RGB images. 
Current foreground extraction techniques are very mature and generalize well. 
We use IS-Net~\cite{qin2022} to perform foreground segmentation in each image. 
The DINO-ViT features can also be used to do foreground segmentation, but we found that IS-NET provides more accurate segmentation masks.

We draw inspiration from BARF~\cite{lin2021barf} and optimize an implicit occupancy field with bundle-adjustments. To do so, we first need to convert the estimated orthographic cameras into perspective cameras.

\begin{figure}
    \centering
  \includegraphics[width=\linewidth]{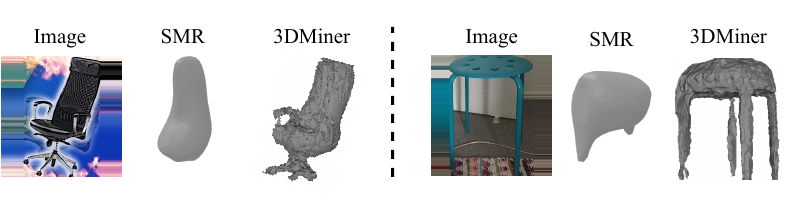}
  \vspace{-8mm}
  \caption{\textbf{Qualitative Comparison on Pix3D chairs.}
  When applied to actual in-the-wild images of objects, 3DMiner generates much more accurate geometry than state-of-art methods like SMR.
  }
  \vspace{-4mm}
  \label{fig:pix3d}
\end{figure}

\paragraph{Orthographic to perspective initialisation.}
For every image, we use the orthographic parameters from $M$ (estimated with Equation \ref{eqn:orthographic}) to initialise a perspective camera-to-world matrix $P$ in  $\mathbb{R}^{4\times 4}$:

\begin{equation}
P = 
    \begin{bmatrix}
    & \frac{m_1}{||m_1||} &  | &\\
    & \frac{m_2 - (p_1\cdot m_2)p_1}{||m_2 - (p_1\cdot m_2)p_1||}  &  T &\\
    &p_1 \times p_2 &  | &\\
    &0 & 1 &
    \end{bmatrix}^{-1},
    \label{eqn:cam_init}
\end{equation}

where $m_i$ and $p_i$ denote the $i^{th}$ row of $M$ and $P$, respectively.
To understand this derivation, recall that the top-left sub-matrix $P_{[1:3, 1:3]}$ is the rotation matrix controlling the camera viewing direction, while $M$, the orthographic projection matrix, is a $ \mathbb{R}^{2\times 3}$ matrix formed by two orthogonal 3D vector and can be interpreted as a linear plane of projection. The rotation corresponding to M is thus simply obtained by applying the Gram-Schimdt orthonormalization process for the first two rows, and getting the cross product for the third, as suggested by \cite{zhou2019continuity}. Note that $M$, which we estimate by Rigid Factorization, is initially orthogonal, so we could simply set  $P_{[1:2, 1:3]}$ to be a normalized version of $M$. However, we seek to optimize $M$ via gradient descent, which does not guaranty that M remains orthonormal throughout the process, hence why Gram-Schimdt orthonormalization is important.
T is a translation vector initialised as $[0, 0, z]^T$, where $z$ is a scalar hyperparameter (set to 5 for our experiments).
We also initialise a camera intrinsic matrix $K$ with focal length $f$ equaling to the image size. This initialization, though  inaccurate, is optimized during the bundle-adjustment.
\paragraph{Bundle-adjusting camera parameters.}
Given $K$ and $P$, we cast rays through each pixels, and sample points along each ray  $r$. Each point  $x$ is encoded with the progressive positional encoding technique from BARF~\cite{lin2021barf} where the $k^{th}$ frequency of the positional encoding is:
\begin{equation}
\gamma(x, \alpha) = w(\alpha)\cdot[\cos 2^k\pi x, \sin 2^k\pi x],
    \label{eqn:pos_encoding}
\end{equation}
where $w$ is a weight controlled by hyperparameter $\alpha$ that gradually increases as the training progresses. This effectively activates the encoding of higher frequencies as training progresses.

We feed the positional encoded inputs into the occupancy field $\mathrm{MLP}$ and obtain an occupancy output. The loss is a  binary cross-entropy  comparing the ground-truth silhouettes occupancy $o_{gt}$ and the soft maximum occupancy of the corresponding ray:
\begin{equation}
\mathcal{L}_r = \mathrm{BCE}(o_{gt}, 1 - e^{(- \sum_{x \in r} \mathrm{MLP}(x, [\gamma(x, \alpha)]_{\alpha=0}^{10}))}).
\label{eqn:loss}
\end{equation}

We jointly optimize the 3D occupancy network, and the bundle-adjustment parameters $f$, $M$, and $T$s for each image. We do not directly optimize the matrix $P$ via gradient descent because it needs to remain in the manifold of rotation matrices.
We follow~\cite{lin2021barf} and use marching cube to extract a mesh from the learned occupancy field.

\paragraph{Regularizing the geometry and space}

For real-world datasets like LAION-5B, the number of images sharing common objects may be very limited. Thus, we draw inspiration from RegNeRF \cite{niemeyer2022regnerf} and impose two extra regularizations during our occupancy field optimization. First, we encourage piece-wise smoothness of objects by imposing an additional geometric regularizer $\mathcal{L}_g$:

\begin{equation}
\mathcal{L}_g = (d(r_{i,j}) - d(r_{i,j+1}))^2 + (d(r_{i,j}) - d(r_{i+1,j}))^2,
\label{eqn:loss_geo}
\end{equation}
where $r_{i,j}$ indicates the ray casted from pixel coordinate $(i, j)$ and $d$ is expected depth calculated in the same manner as \cite{niemeyer2022regnerf}.
Second, since all our reconstructed shapes can be placed at the center of the coordinate system, we impose space annealing to confine the near and far plane then gradually expanding it as training iteration progresses.
\begin{figure}[t]
    \centering
  \includegraphics[width=\linewidth]{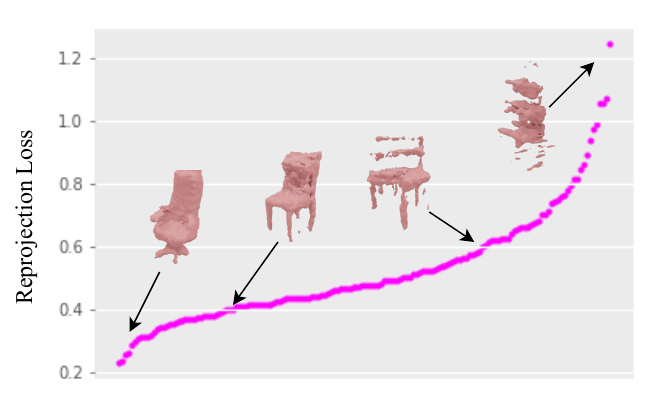}
  \caption{\textbf{Reprojection error.} We plot the reprojection error per cluster (averaged over each image in the cluster) in ascending order and show representative reconstructions for four data points. We empirically observe that the reprojection error is a good indicator of the quality of the reconstruction. 
  }
  \label{fig:threshold}
\end{figure}


\section{Experiments}
\begin{figure*}[t]
\centering
\includegraphics[width=\linewidth]{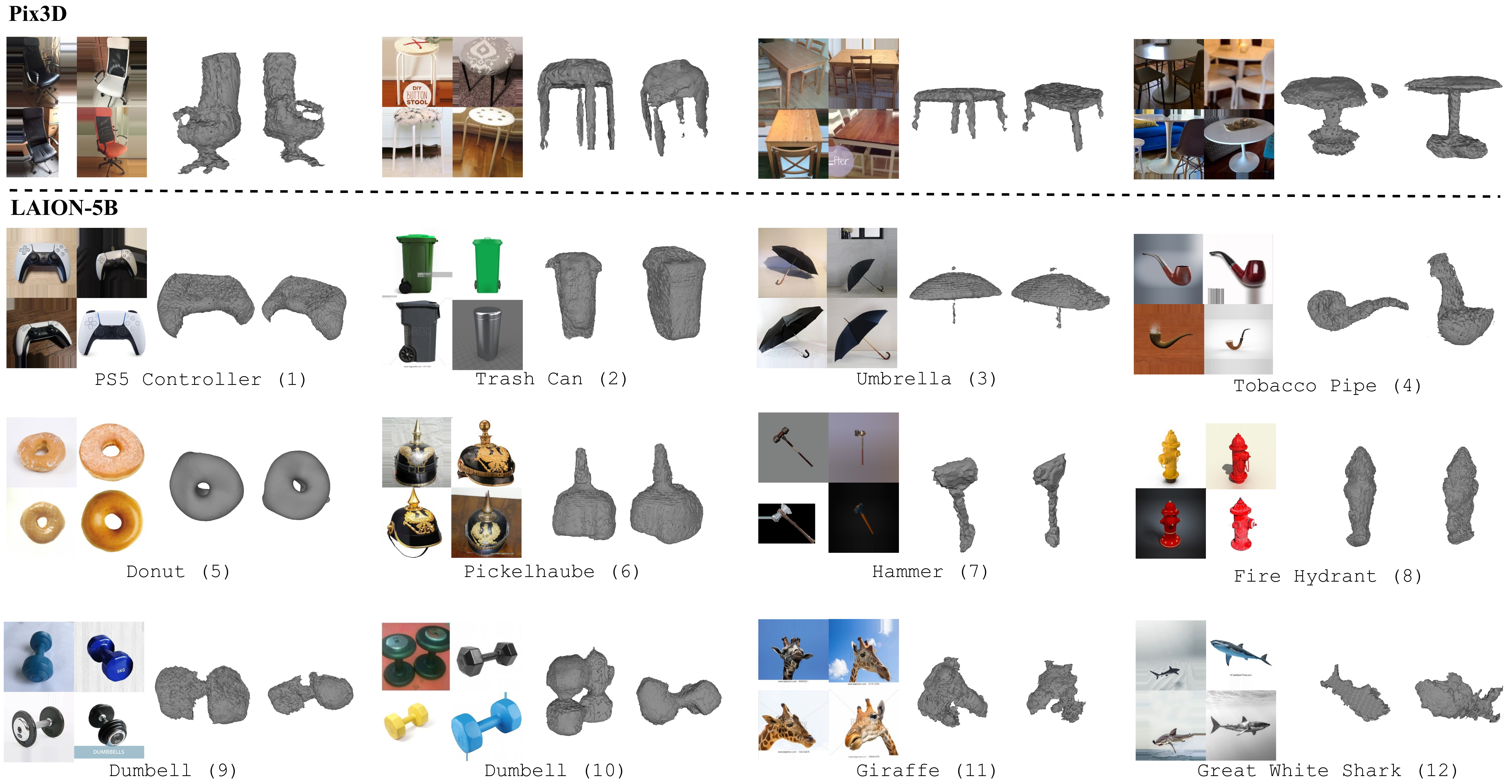}
\caption{\textbf{Qualitative Results on Pix3D and LAION-5B.}  We show examples of mining 3D shape from the in-the-wild LAION-5B dataset with the above text prompts. \texttt{1}$\sim$\texttt{6} show reconstructions with high fidelity when clusters are very accurate. \texttt{7}$\sim$\texttt{10} present generic shapes captured when the clusters are more diverse. \texttt{11}$\sim$\texttt{12} are results on very challenging non-rigid objects.}
\vspace{-4mm}
\label{fig:laion}
\end{figure*}
\subsection{Implementation details}
All our models are trained on a single Tesla V100 machine. We use 10  augmentations for our image clustering step. For keypoints, we use the 8 most up-voted segments unless specified. We sample 32 rays and 32 points on each ray during every iteration and train the model for 300 epochs with $\alpha$ increasing from 1 every 20 epoch up to 10, using an Adam optimizer with a learning rates of $10^{-3}$. The learning rate for camera parameters decay by a factor of 0.1 every 100 epochs. 

\subsection{Comparison on Pix3D chairs}

\label{sec:pix3d}

To validate our approach, we directly train the model on the very challenging Pix3D chairs dataset~\cite{sun2018pix3d} consisting in 3839 real images of  561 different  chairs. This is the most difficult Pix3D category and has been used by prior work as the benchmark \cite{tars3d,xie2020pix2vox++} (please see Appendix for additional Pix3D categories). We compare against two state-of-the-art approaches: SMR~\cite{smr} and Unicorn~\cite{monnier2022share}.  
Since our method is completely unsupervised and consists of an optimization for every cluster of images, there is no learning involved. 
Hence, there is no need to split a training and testing dataset. 
We retrain SMR~\cite{smr} and Unicorn~\cite{monnier2022share} on the entire set of images, and evaluate them on the same set. Therefore, all three methods receive the same amount of information during weight optimization.

To evaluate 3DMiner, for each image, we query which cluster it belongs to, and map the image to the corresponding 3D reconstruction. 
For all methods, we compare the 3D reconstruction in a canonical orientation against their ground truth 3D shape, 
and average performances across all input images. 
To focus the evaluation on the quality of the geometries, we align the reconstructions to their ground truth with Coherent Point Drift~\cite{coherentpointdrift}, optimizing for translation, rotation and uniform scaling. 
SMR requires image masks, while Unicorn only needs an image. 
To make the comparison fair, we also use masked images for Unicorn and use ground truth masks instead of estimated masks in the last step of our pipeline.

As shown in Table \ref{tab:pix3d}, our approach greatly outperforms SMR and Unicorn, reducing the Chamfer Distance by 32\% and improving the F-score by 10 points. 
We found that training Unicorn led to a degenerate solution where the network always predict the same mean shape. 
This aligns with the author's feedback on the official GitHub implementation \footnote{https://github.com/monniert/unicorn}, suggesting that the model is very difficult to train on real-world datasets. By contrast, our meshes are instance-specific and therefore more accurate on a variety of chairs. In Figure~\ref{fig:pix3d}, we provide a short qualitative comparison of 3DMiner against SMR (please see Appendix for more).
Additional results for our method in chairs and tables are also presented in Figure \ref{fig:laion}.
\paragraph{Filtering 3D shapes.}

To mine 3D data from images, it is crucial to automate the process of distinguishing good reconstructions from bad ones. We hypothesize that the reprojection error, which is the final converged loss of our bundle-adjusting reconstruction (see Equation~\ref{eqn:loss}), correlates with the reconstruction quality.
To validate this observation, we show a couple of experiments.
First, for each cluster in Pix3D Chairs, we plot the reprojection error averaged per image and show representative rendered meshes in Figure \ref{fig:threshold}. 
An error smaller than 0.4 generally indicates good consistency across all views (22\% of the Pix3D clusters). 
This consistency drops significantly as the error increases, and degenerate solutions generally appear  when the error goes over 0.8 (8\% of the Pix3D clusters). 
Second, in the quantitative study on Pix3D in Table~\ref{tab:pix3d}, we select images whose cluster has a lower reprojection error than a certain threshold, and average the reconstruction error only on this subset of images. The results show that the reconstruction quality improves significantly as we decrease the threshold. As reference, we also present the results of other baselines on the same subset of images.
    
\subsection{In-the-wild dataset: LAION-5B}   

\begin{figure}[t]
    \centering
  \includegraphics[width=0.8\linewidth]{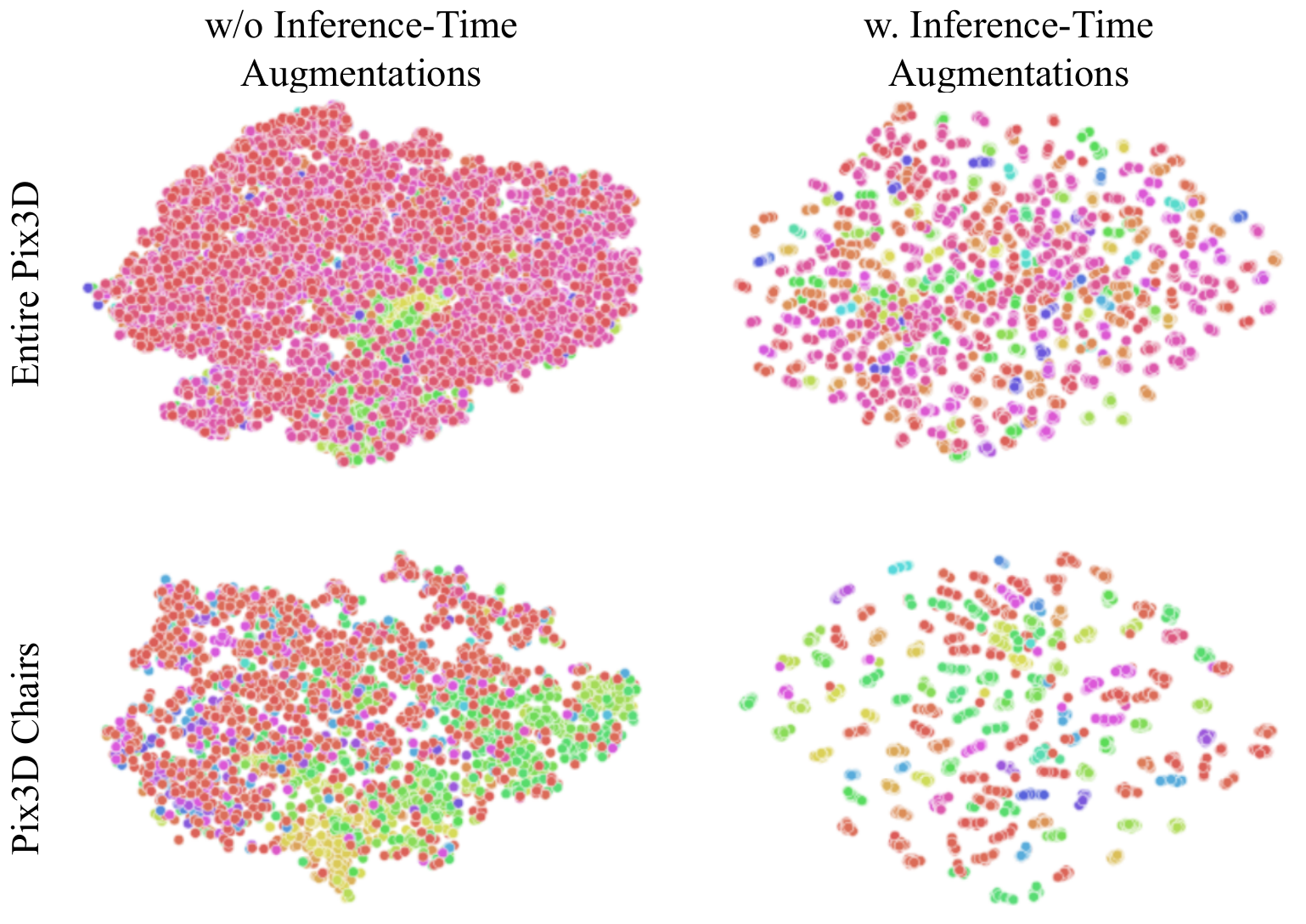}
  \caption{\textbf{Ablation of our augmentation scheme.} We show the T-SNE embeddings of Pix3D images embedded with DINO-ViT~\cite{caron2021emerging} with and without our augmentations scheme.  Each color denotes a different object. 
  Note that with augmentations, images corresponding to the same 3D object are grouped together, forming clusters on a single color (right), which is not the case without augmentations (left).}
  \label{fig:inf-augmentation}
\end{figure}

To showcase the capability of our 3DMiner, we present, to the best of our knowledge, the first 3D reconstruction results from images in LAION-5B. 
Lower part of Figure~\ref{fig:laion} shows our results on 12 categories. 
We download the first 500 images returned from LAION-5B using various text prompts \footnote{https://rom1504.github.io/clip-retrieval/}, then run 3DMiner on the resulting datasets, with varying distance thresholds (\textit{i.e.,} different numbers of images within the clusters). 
We filter out the reconstructions with reprojection error $> 0.4$ per the analysis in the previous section. 

As we can see in Figure~\ref{fig:laion}, the image from LAION-5B are noisy, but our clustering leads to cleaner subsets of images. 
Specifically, reconstructions \texttt{1}$\sim$\texttt{6} shows reconstructions with high fidelity as clusters are generally very clean, even when the color and backgrounds vary drastically. 
Intriguingly, when the clusters are less precise (\textit{e.g.,} reconstructions
\texttt{7}$\sim$\texttt{10}), 3DMiner still captures the generic shape. In particular, \texttt{10} clustered images of one and two dumbells, but the bundle-adjustment allows us to find an angle where both images are valid.
Finally, we also investigated how the method would perform for non-rigid objects (\texttt{11}$\sim$\texttt{12}). As expected, our method has a lot of trouble dealing
with non-rigid shapes, but ends up reconstructing almost-rigid portions of the geometry (e.g head of the giraffe).

\noindent\textbf{Failure cases on LAION-5B.} 3DMiner fails on  several text prompts. We identified the two sources of failures:
\textit{(i) Not Enough Angles.} Prompts like \texttt{fish} exhibit only the side  and not the front/back view. This makes it hard to estimate the depth of the object and update the focal length and camera translation accordingly. 
\textit{(ii) Watermarks} appear in a large portion of LAION-5B data. 
Our estimated masks accidentally capture the watermarks as salient objects, 
which severely undermines our reconstruction results. 
Failure prompts include \texttt{Statue of Liberty}.

    
\subsection{Analysis}

\label{sec:inf-augmentation}

\begin{table}[t!]
    \centering
        \resizebox{0.65\linewidth}{!}{
    \begin{tabular}{lcc}
    \toprule
     \textbf{Method} & \textbf{Dist. Thres.}&\textbf{NMI}\\
     \cmidrule(lr){2-3}
            Original Image & 10  &0.647 \\
            Image + Aug    & 10 & 0.758\\
            Image + Aug    & 30 & 0.786\\
            Image + Aug    & 100& 0.764\\
    \bottomrule
    \end{tabular}
    }
    \caption{\textbf{Normalised mutual information comparisons}. We compare the Normalised Mutual Information (NMI) of the ground-truth labels against our predicted clusters. ``Aug'' denotes our augmentations scheme (see Section~\ref{sec:cluster}), and ``Dist. Thres.'' denotes the threshold distance set for agglomerative clustering. Our augmentation scheme improves the quality of the cluster (+11 points).}
    \label{tab:inf-augmentation}
    \end{table}
    
\noindent\textbf{Augmentations before image clustering.}
We validate the contribution of our augmentation scheme on Pix3D images. 
Figure \ref{fig:inf-augmentation} shows a T-SNE visualization of the embedding space created with and without the augmentation. 
The colors denote the ground truth clusters. 
Our augmentation scheme clearly helps the DINO-ViT features to be more instance-specific, even when the images come from multiple categories. 
We further quantify the quality of our clusters by measuring the normalised mutual information (NMI) with the ground-truth clusters. Table \ref{tab:inf-augmentation} shows that adding augmentations improves the NMI by 0.11, and that the quality of the clusters remain stable when we vary the distance threshold used during agglomerative clustering. 
This is a useful feature when applying 3Dminer to real-world datasets: we can automatically test several distance thresholds to gather enough images within each cluster while maintaining cluster precision.

\begin{figure}
    \centering
  \includegraphics[width=\linewidth]{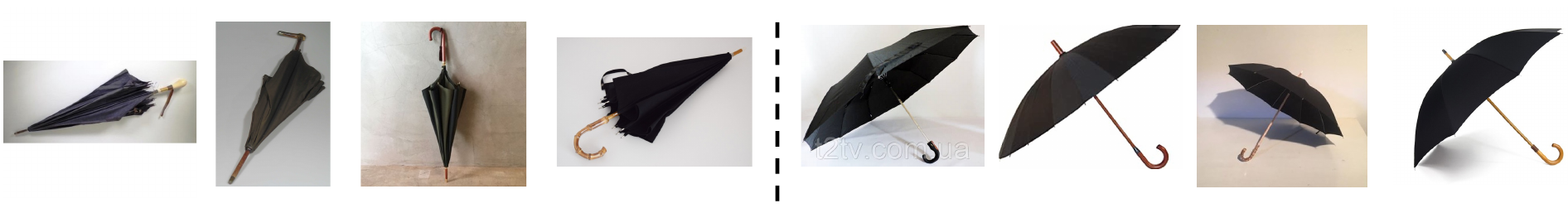}

  \caption{Two clusters from text prompt \texttt{Umbrella}, showing non-rigid objects separated based on part-wise poses.}
  \vspace{-3mm}
  \label{fig:non-rigid}
\end{figure}

\noindent\textbf{Clustering Non-rigid Objects.} One additional benefit of doing the clustering is that it tends to group objects where the part-wise poses are also similar. We show in Figure \ref{fig:non-rigid} an example taken from the prompt \texttt{Umbrella}, where opened and closed umbrellas are grouped into different clusters, allowing us to reconstruct the object even with certain level of non-rigidity (The right cluster is what allows the reconstruction of umbrella shown in Figure \ref{fig:laion}). Tightening the clustering threshold would also constrain rigidity within a cluster, though with the trade off of fewer images and thus a higher likeliness of degenerate 3D reconstructions (\textit{e.g.}, most cases of clusters with only 3 or 4 images tend to fail).

\begin{figure}[t]
    \centering
  \includegraphics[width=0.9\linewidth]{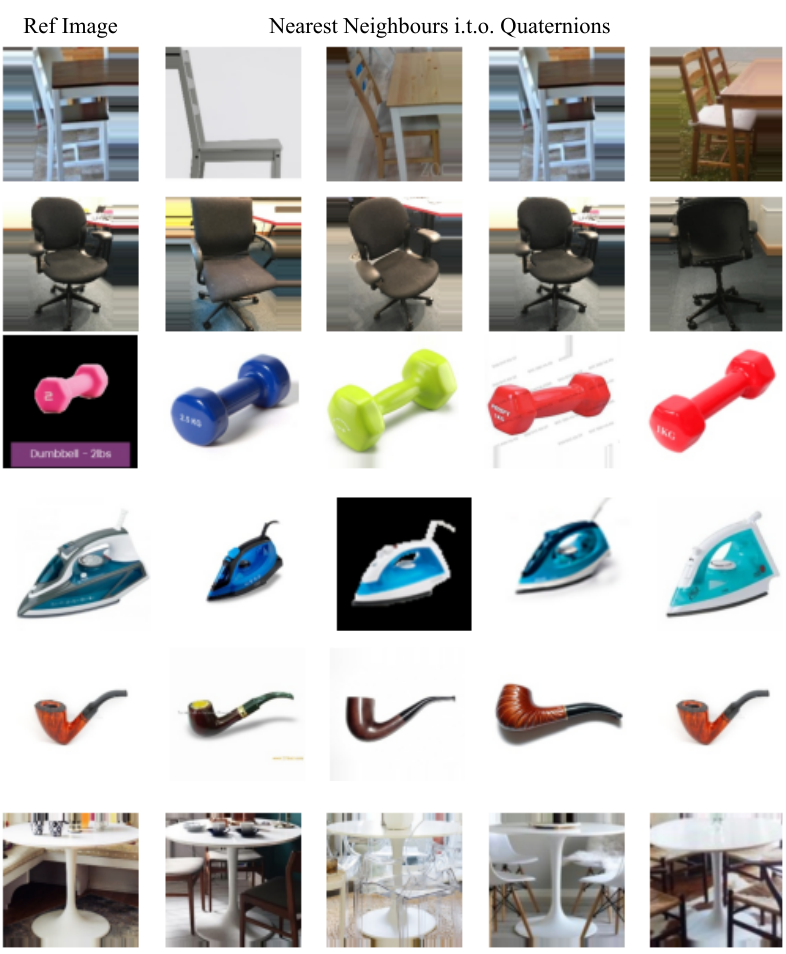}
  \vspace{-2mm}
  \caption{\textbf{Pose estimation Analysis.} We randomly select a reference image (leftmost column) out of a cluster and find the images with the nearest quaternion poses (rightmost 4 columns).}
  \label{fig:pose_analysis}
\end{figure}

\vspace{1mm}\noindent\textbf{Pose Estimation on Similar Objects.}
In addition to shapes, 3DMiner also provides association images and \emph{posed shapes} -- shapes that are aligned with the image content when the estimated pose is applied. We show a qualitative evaluation in Figure~\ref{fig:pose_analysis}. 
Specifically, given a cluster, we take a random image for reference (leftmost column), and find the 4 nearest neighbours in terms of their quaternion poses within the cluster (right 4 columns). We show that even when the shape, texture, and background differs, our bundle-adjustment still allows us to capture a rough pose among them leading to promising shape reconstructions.


\section{Conclusion}

We have presented 3DMiner, a novel pipeline for mining 3D shapes from large-scale unnanotated in-the-wild image datasets. 
Differently from single network end-to-end approaches,
our technique can be thought of as a reincarnation of classical approaches~\cite{reconstructingpascalvoc}
while replacing manual annotations with representations learned from deep networks.
The key elements of our pipeline are: \textit{(i)} a clustering step using DINO-ViT features, \textit{(ii)} a camera estimation step using a classical Structure-from-Motion technique and keypoint estimation, and \textit{(iii)} a progressive bundle adjusting reconstruction to learn an occupancy field supervised by image silhouettes.
Through rigorous experiments, we have showed that 3DMiner outperforms the state-of-the-art on the Pix3D dataset, 
and to the best of our knowledge, is the first to show 3D reconstruction results on the LAION-5B dataset. 
We hope the 3DMiner serves as a testbed for a newly proposed task of mining geometry from large-scale unannotated datasets and all the subproblems it involves.

\noindent\textbf{Limitations.}
Although our model is able to reconstruct 3D shapes solely from in-the-wild images, the concavity of the shapes is not captured. 
This is  because the occupancy field is supervised with a silhouette loss, which amounts to space carving. 
Future works could explore using monocular depth estimation networks as further supervision in the reconstruction problem. 
Furthermore, 3DMiner is a sequential pipeline and thus vulnerable to the failure of any step.
Fortunately, we verified that we can automatically identify bad 3D reconstructions in order to discover only
reasonable 3D shapes.
As every component of the pipeline eventually advances, we hope that the number of meaningful 3D shapes discovered
from image datasets can be increased.
Finally, we also hope to investigate the utilization of using 3DMiner generated shapes for supervision in training better SVR techniques.

\bigbreak
\noindent\textbf{Acknowledgements.} This work is supported in part by the EPSRC ACE-OPS grant EP/S030832/1.

{\small
\bibliographystyle{ieee_fullname}
\bibliography{egbib}
}

 \pagebreak

\section*{Supplementary Material}
In this supplementary material, we provide additional  comparisons on Pix3D tables and sofa, as well as qualitative visualizations on every step of our 3DMiner pipeline with a Pix3D and a LAION example.

\section{Additional Comparisons}
On top of the evaluations on Pix3D chairs in the paper, we also provide additional results on Pix3D tables and sofas. Additionally, we evaluate the effectiveness of bundle-adjustment on the reprojections and overall reconstruction quality.
\subsection{Quantitative Comparisons}
\begin{table}[t!]
    \centering
        \resizebox{0.65\linewidth}{!}{
    \begin{tabular}{lcc|cc}
    \toprule
    &\multicolumn{2}{c|}{Table} &\multicolumn{2}{c}{Sofa} \\
    & \textbf{CD$\downarrow$}&\textbf{F1$\uparrow$}& \textbf{CD$\downarrow$}& \textbf{F1$\uparrow$}\\
     \cmidrule(lr){2-3} \cmidrule(lr){4-5}
            SMR      & 0.226 &0.063 &0.165 & 0.092\\
            Unicorn  & 0.328 &0.035 &0.120 & 0.122\\
            3DMiner  & \textbf{0.180} &\textbf{0.103} &\textbf{0.097} & \textbf{0.168}\\
    \bottomrule
    \end{tabular}
    }
    \vspace{2mm}
    \caption{\textbf{Comparisons on Pix3D Tables and Sofas}. We use Chamfer Distance (CD) and F1 Score (F1) as the standard metrics. Due to computational restrictions, we are not applying all the regularizations for this table and the table in the main paper.}
    \label{tab:inf-augmentation}
    \end{table}

We provide the Chamfer distance (CD) and F1 score (F1) comparisons of our 3DMiner against SMR and Unicorn. We can see a clear improvement on both categories (over 40\%).

\subsection{Qualitative Comparisons}
\begin{figure}[t]
\centering
\includegraphics[width=\linewidth]{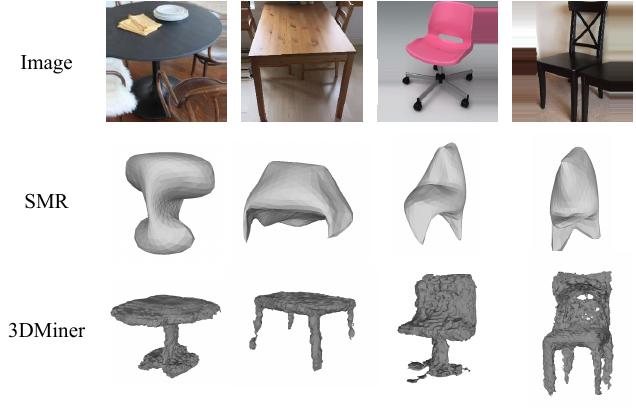}
\caption{\textbf{Quantitative Comparisons on Pix3D Chairs and Tables}. We show the reconstruction results of four instances against SMR.}
\vspace{2mm}
\label{fig:raw_imgs}
\end{figure}
We also juxtapose the reconstructions of 3DMiner and SMR. Note that Unicorn fails to perform any meaningful reconstrutions (all blocks) due to the very challenging scenes.

\begin{figure}[t]
\centering
\includegraphics[width=\linewidth]{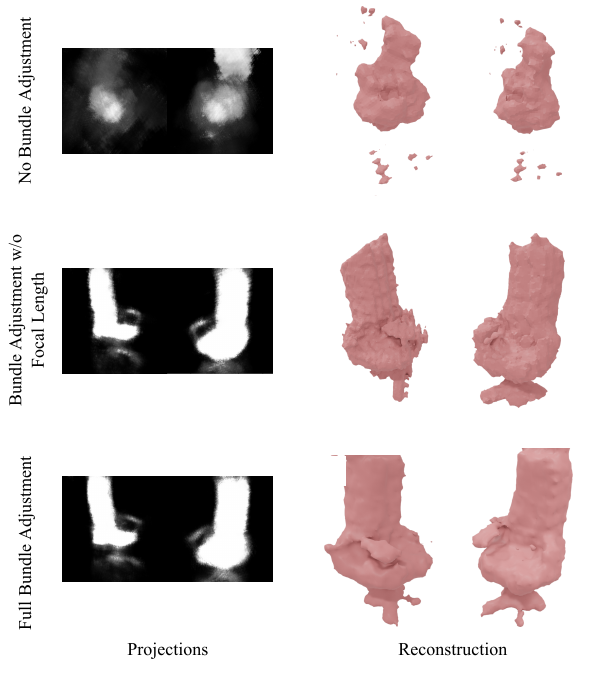}
\caption{\textbf{Ablations on Bundle-adjusting poses and focal lengths}. We visualize the results of bundle adjustment on the reprojection error and overall reconstruction quality}
\label{fig:ba-ablation}
\end{figure}

\begin{figure}[t]
\centering
\includegraphics[width=\linewidth]{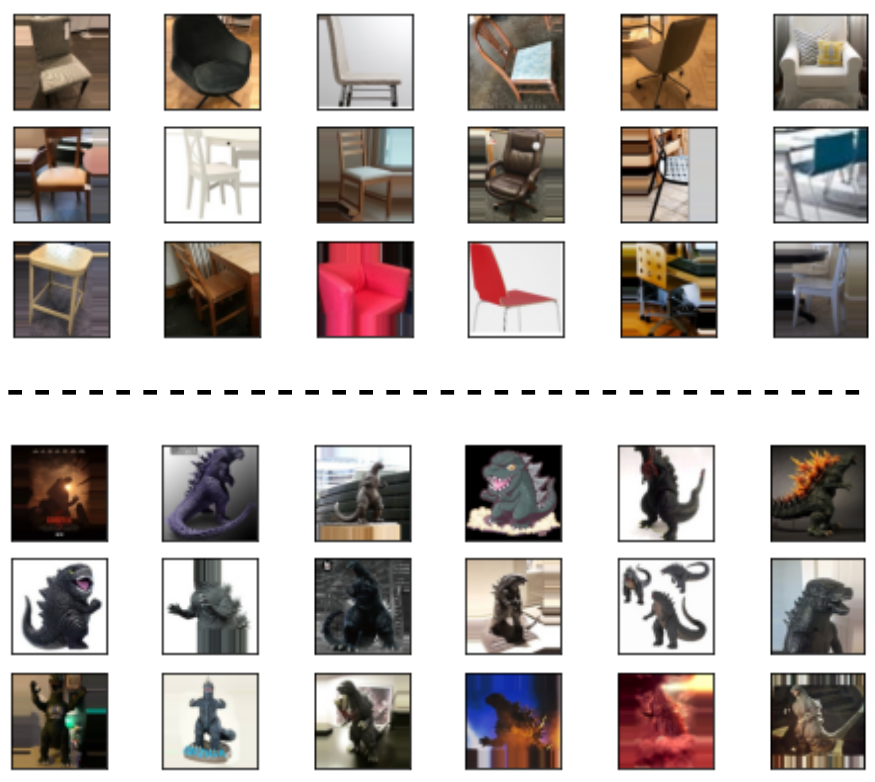}
\caption{\textbf{Samples of the Original Image Set}. Image above the dotted line are from Pix3D and below are from LAION-5B Godzilla. You can easily spot the noisiness within the original datasets.}
\label{fig:raw_imgs}
\end{figure}

\subsection{Ablations on Bundle-adjustment}

We hope to understand the capability of our orthographic-to-perspective bundle adjustment. To rule out any uncertainties we select one Pix3D cluster that is $100\%$ accurate (\textit{i.e.,} all images depict the same 3D shape). Our reconstruction achieves roughly $0.38$ F-Score. We present the reconstructed meshes under three training settings: \textit{ (i)} no bundle-adjustment at all, \textit{(ii)} bundle adjusting only the motion and translation, and \textit{(iii)} bundle adjusting all camera parameters together.

As shown in Figure \ref{fig:ba-ablation}, even though the initial rigid factorization algorithm constrains all cameras to point to the same shape, the poses is very rough in the first place. This leads to a very coarse and degenerate solution (without any handles and legs). As we introduce motion and translation, we gain significantly more details of the chair. Nevertheless, due to the focal length inaccuracy, we may end up with artifacts (\textit{e.g.,} only one armrest, missing legs) or even rougher edges due to inconsistencies between the projections. Learning the focal lengths is thus important to recover details in the geometry.

\begin{figure}[t]
\centering
\includegraphics[width=\linewidth]{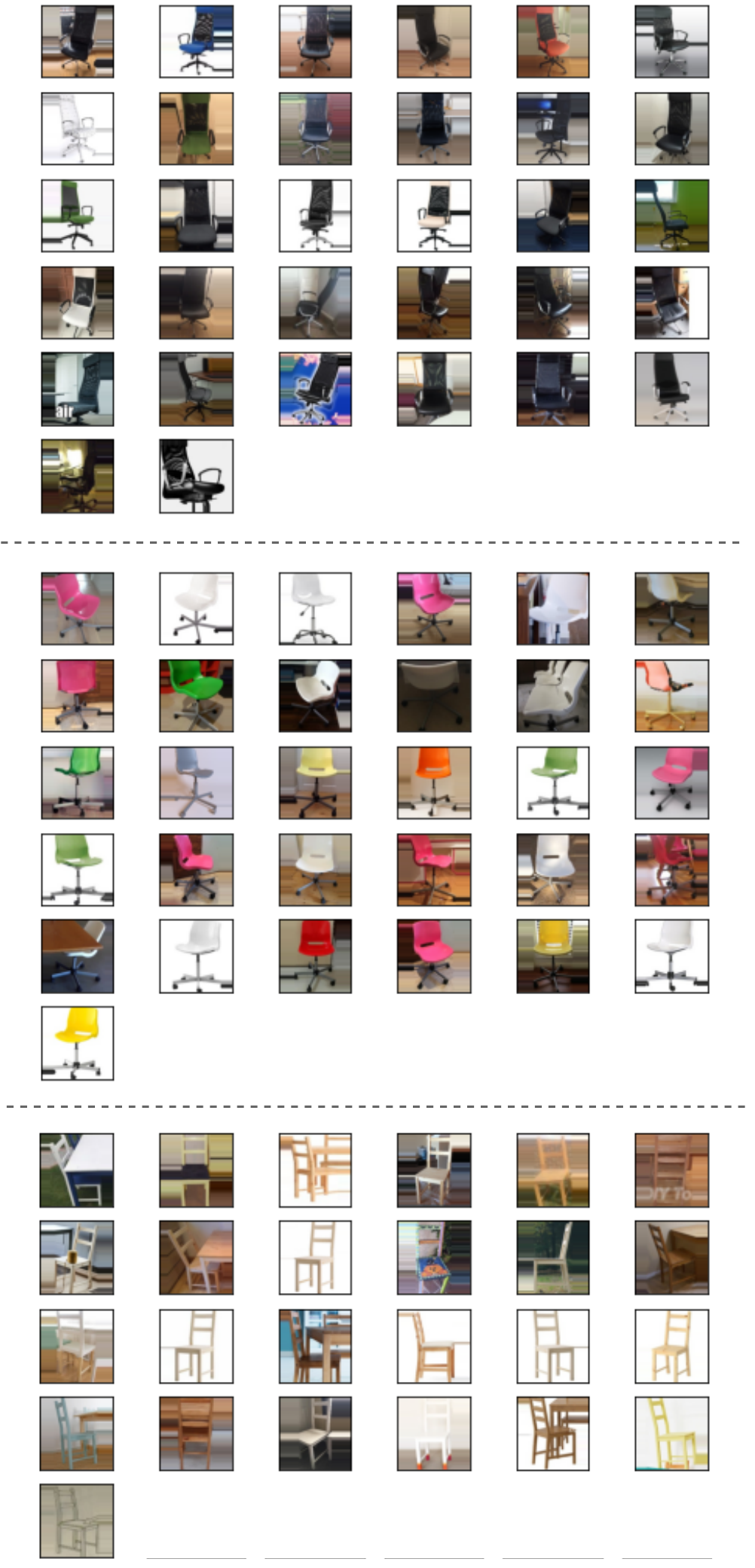}
\caption{\textbf{Example Clusters from Pix3D.}
Each set of images separated by dotted lines shows an image cluster computed from Pix3D chairs
using the method described in the main paper in Section 3.1.
Notice how even when the have objects in different colors, textures, completely different backgrounds and illumination,
one set of images still roughly depicts objects with the same geometry.}
\label{fig:cluster_pix3d}
\end{figure}

\begin{figure}[t]
\centering
\includegraphics[width=\linewidth]{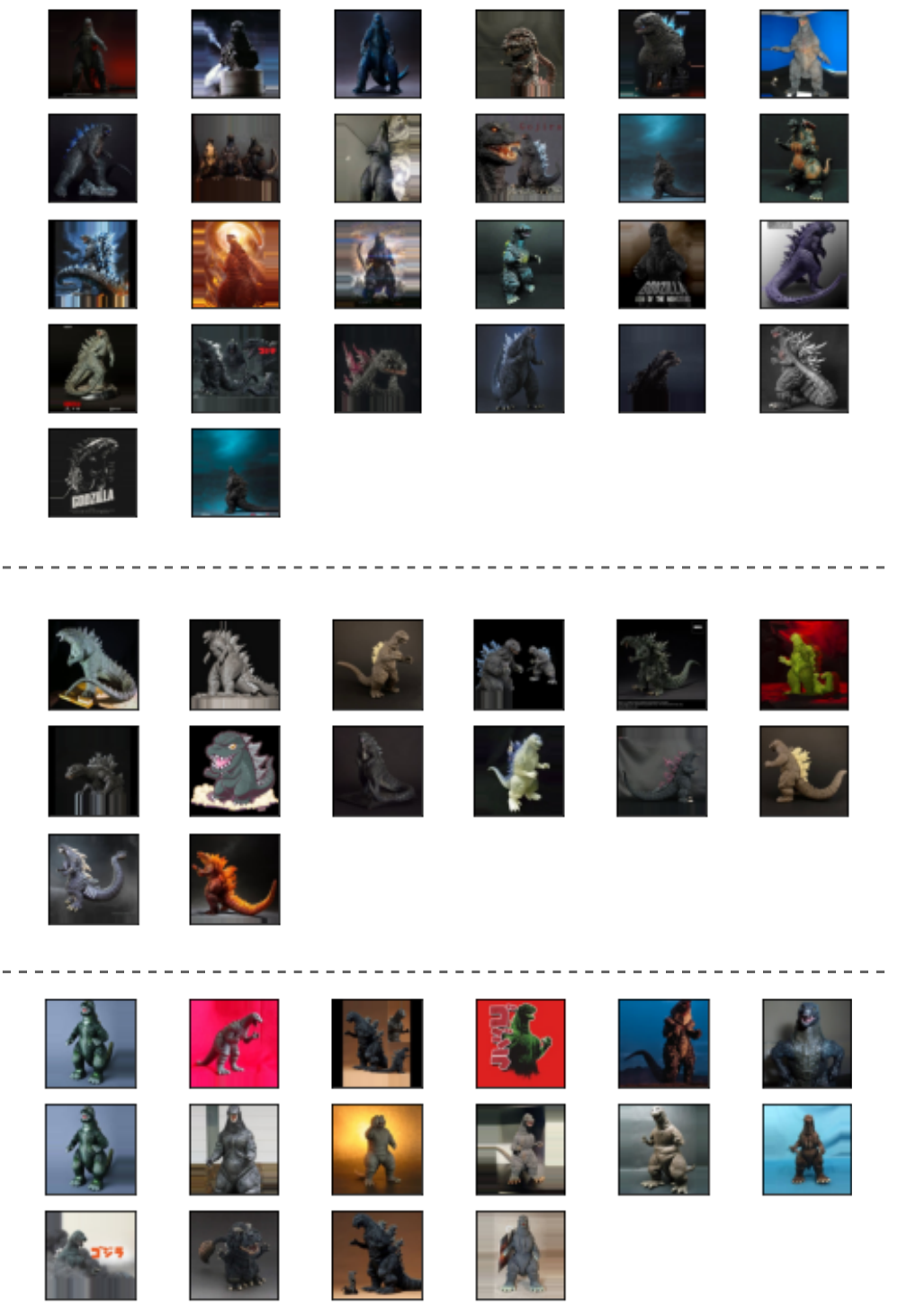}
\caption{\textbf{Example Clusters from LAION-5B.}
Each set of images separated by dotted lines shows an image cluster computed from LAION-5B using the prompt
\texttt{Godzilla}.
Clusters were computed using the method described in the main paper in Section 3.1.
}
\label{fig:cluster_laion}
\end{figure}

\section{Clustering Images with Similar Objects}
This corresponds to applying the method described in Section 3.1 of the main paper.
The main idea of this experiment is to show how the clustering procedure groups
images depicting similar shapes together, even when presented with heavy nuisance factors --
objects have different colors, materials, textures and in completely different environments. As a reference, we present subsets of the unclustered images in Figure \ref{fig:raw_imgs}
As we can see, creating representations through image augmentations followed by feature pooling leads to representations that
are robust to these nuisance factors and mostly encode shape information.
We show three examples each of clusters from the two datasets, Pix3D and LAION-5B \texttt{Godzilla}, in Figure~\ref{fig:cluster_pix3d} and Figure~\ref{fig:cluster_laion}, respectively. 

\section{Initial Pose Estimation}
This section illustrates the method described in Section 3.2 of the main paper.
It is mainly divided in two parts.
The first one analyzes local image features from all the images groups their pixels according to
feature similarity.
This establishes correspondences between pixels of different images.
The second step uses the aforementioned correspondences in a rigid factorization procedure and yields an
orthographic camera pose for every image within the cluster.

\subsection{Part Segmentation and Masks}

\begin{figure}[ht]
\centering
\includegraphics[width=1\linewidth]{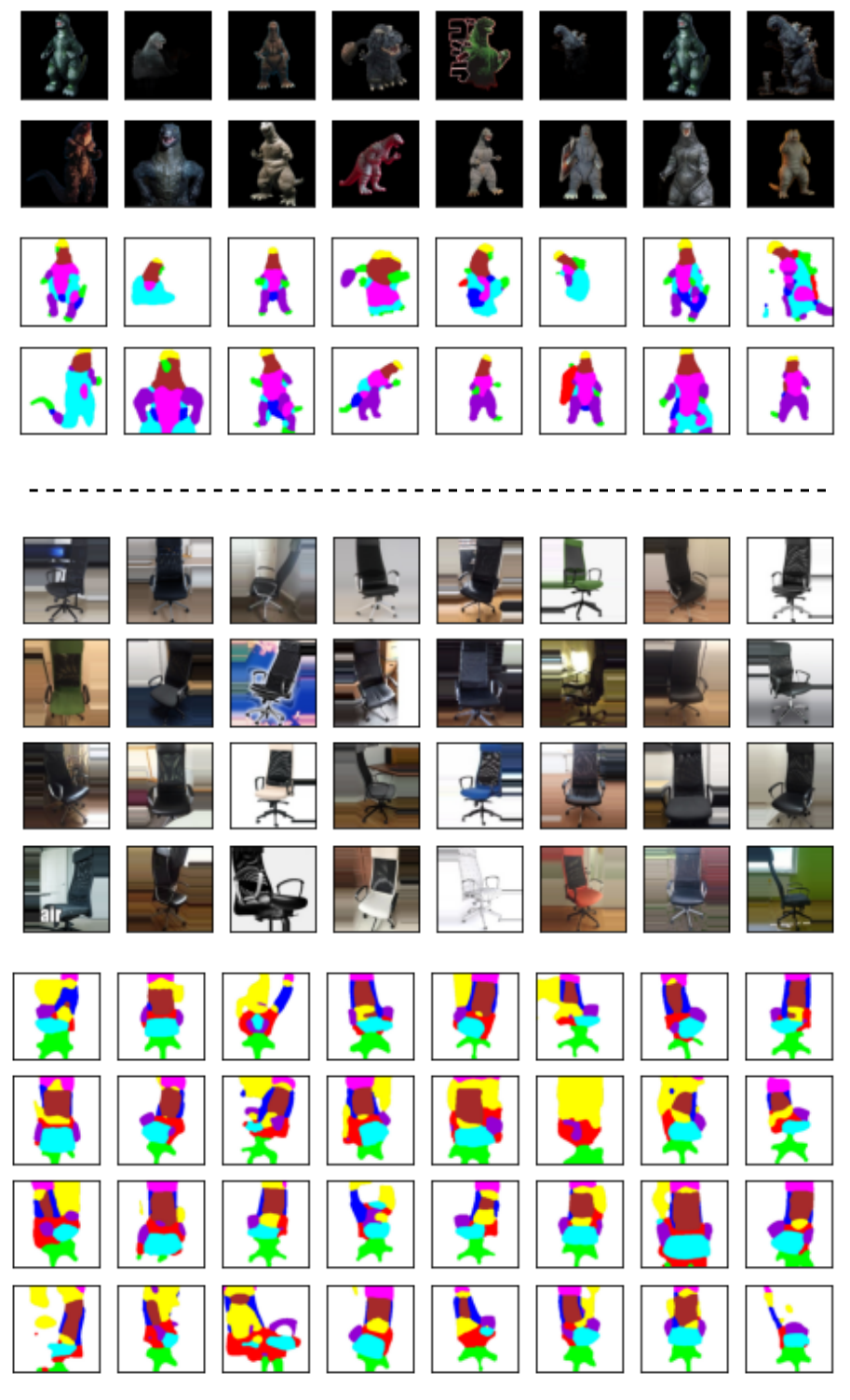}
\caption{\textbf{Part Segmentations.} We visualize the part segmentations of a Pix3D chair and a LAION-5B Godzilla cluster. Note that the Godzilla images goes through ISNet for background removal (First two columns) before the part segmentation. We realized that the Pix3D part segmentations work better with background.
Notice that the part segmentations are far from perfect but their overall semantics is consistent across all the images in the cluster.
For example, all the office chairs have the wheels labeled with green parts, whereas most of the godzillas have their legs labeled with
purple parts, brown heads, and so on.
}
\label{fig:part_seg_masks}
\end{figure}

\begin{figure*}[t]
\centering
\includegraphics[width=0.8\linewidth]{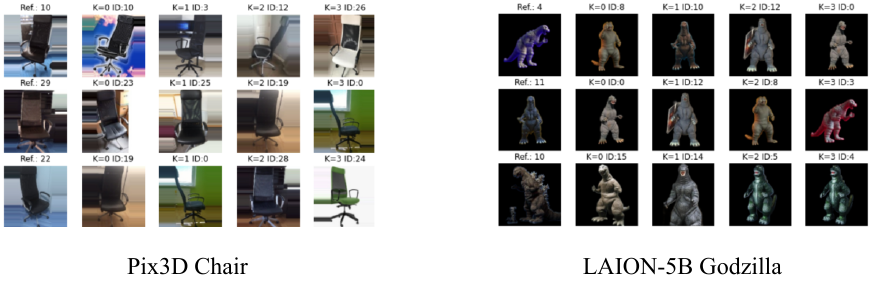}
\caption{\textbf{Orthographic Pose Initialization.} We select 3 anchor image (left column) and output the set of images that are in nearest quarternion distance with the anchors.}
\label{fig:rigid_pose}
\end{figure*}

We present part segmentations for both Pix3D and LAION-5B in Figure \ref{fig:part_seg_masks}. \textbf{The center point of the part segmentations are the keypoints}. The part segmentation is performed by extracting the dense features from DINO-ViT then finding common features across images by K-means clustering.
Note that the first two columns depicting the LAION cluster have already gone through ISNet for background removal 
(for Pix3D we use ground truth masks). 
We observed that keeping the background yielding visually more appealing part segmentations for Pix3D 
especially under the cases of occlusion.

\subsection{Orthographic Pose Estimation} 

\begin{figure}[ht]
\centering
\includegraphics[width=\linewidth]{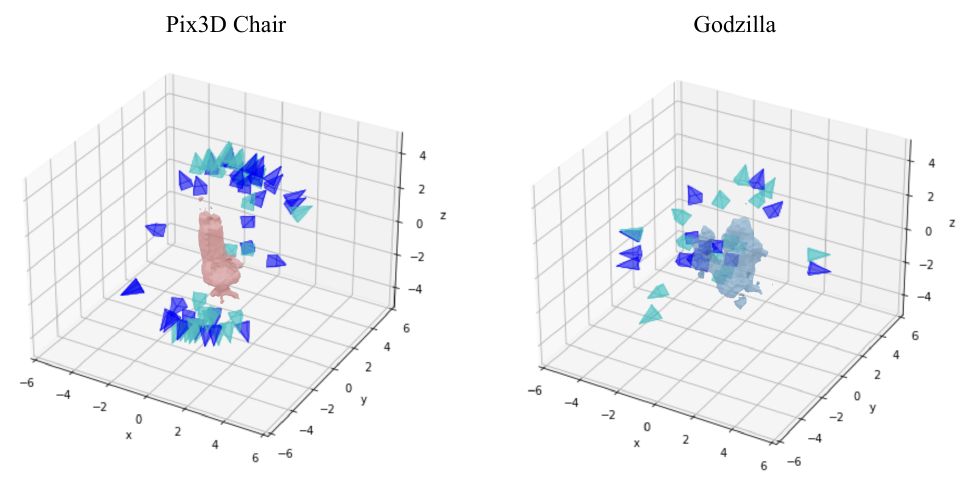}
\caption{\textbf{Camera Poses after Bundle Adjustment}. We visualize the cameras before (blue) and after bundle adjustment (cyan).}
\label{fig:cam_poses}
\end{figure}
We use the previously computed correspondences to estimate orthographic camera poses using rigid factorization to obtain a reasonable initial pose, illustrated in Figure \ref{fig:rigid_pose}.
For this image, find three random images per cluster and present the 4 nearest neighbors in terms of their quaternion distance. 
We can see that, even though the estimations are far from perfect, the poses serve as good initial points to the orientation of the object.

\section{Shape Estimation \& Bundle Adjustment }

\begin{figure}[ht]
\centering
\includegraphics[width=1.0\linewidth]{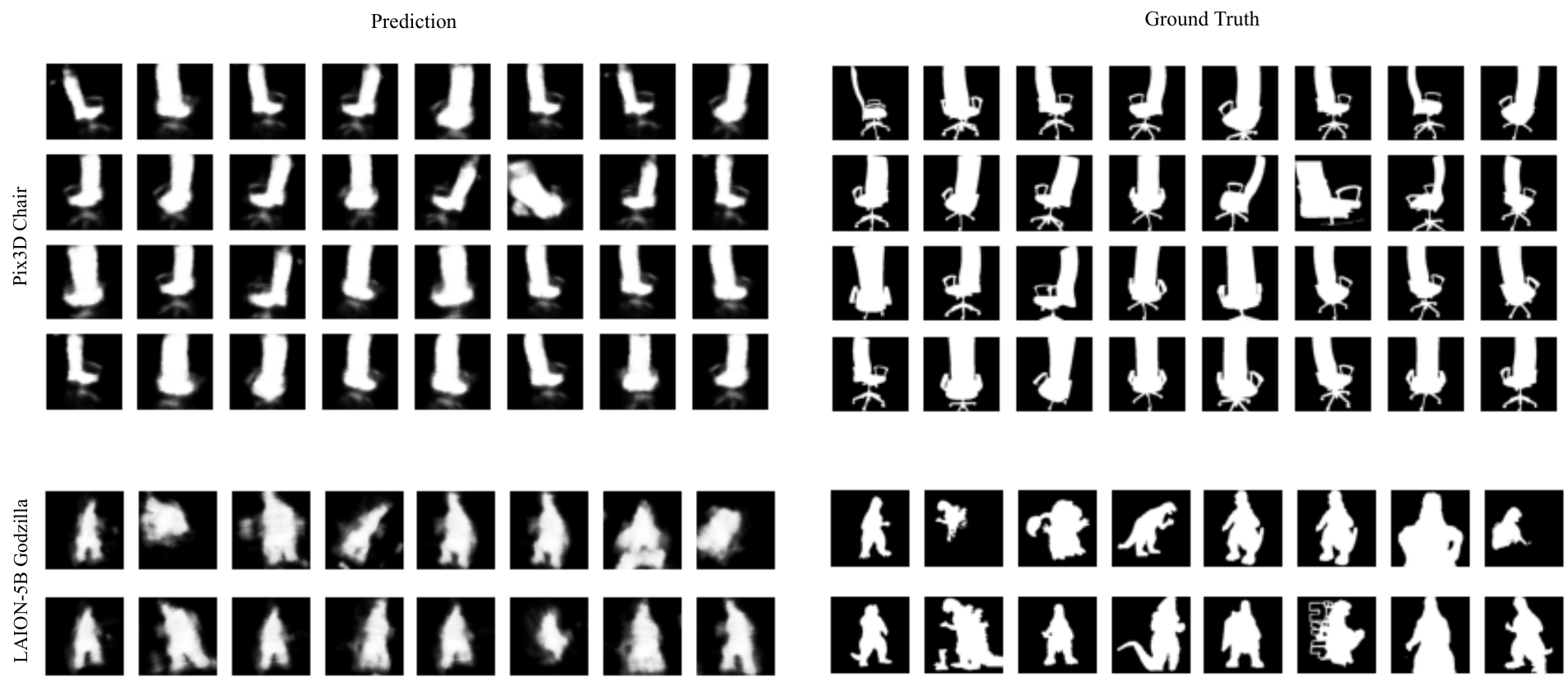}
\caption{\textbf{Reprojections.} We visualize the reprojections against the ground truth for the Pix3D chair and LAION-5B Godzilla.}
\label{fig:reproj}
\end{figure}

Finally, we show the quality of our occupancy field through Figure \ref{fig:reproj} and \ref{fig:cam_poses}. Figure \ref{fig:reproj} shows the reprojection errors after training, and Figure \ref{fig:cam_poses} depicts the changes in the camera poses through bundle adjustment.

\end{document}